\documentclass{article}

\PassOptionsToPackage{numbers, compress}{natbib}
\usepackage[final]{nips_2017}

\usepackage{amsmath}
\usepackage{pgf}
\usepackage{amsfonts}
\usepackage[utf8]{inputenc} 
\usepackage[T1]{fontenc}    
\usepackage{hyperref}       
\usepackage{url}            
\usepackage{booktabs}       
\usepackage{amsfonts}       
\usepackage{nicefrac}       
\usepackage{microtype}      

\title{High performance ultra-low-precision convolutions on mobile devices}

\author{
  Andrew Tulloch and Yangqing Jia \\
  Facebook\\
  \texttt{\{tulloch,jiayq\}@fb.com} \\
}

\begin{document}

\maketitle

\begin{abstract}
  Many applications of mobile deep learning, especially real-time computer
  vision workloads, are constrained by computation power. This is particularly
  true for workloads running on older consumer phones, where a typical device
  might be powered by a single- or dual-core ARMv7 CPU. We provide an
  open-source implementation and a comprehensive analysis of (to our knowledge)
  the state of the art ultra-low-precision (<4 bit precision) implementation
  of the core primitives required for modern deep learning workloads on ARMv7
  devices, and demonstrate speedups of 4x-20x over our additional
  state-of-the-art \texttt{float32} and \texttt{int8} baselines.
\end{abstract}

\section{Introduction}

Recent years have ignited a large interest in low-precision approximation
methods for training and evaluating neural networks --- especially deep
convolutional networks used in computer vision tasks. These methods typically
focus on one (or both) of two objectives --- reducing model size (for example, to
reduce the time spent sending a model over a network to a client), or to improve
performance (by reducing memory bandwidth, using more efficient ALUs, etc). Some
examples of this work include XNOR-Net \citep{Rastegari:2016aa}, QNNs
\citep{Hubara:2016aa}, DoReFa-Net \citep{Zhou:2016aa},
HWGQ-Net \citep{Cai:2017aa}, and several others.

One area of particular interest is the case where both activations and weights
are both uniformly quantized --- as in XNOR-Net, HWGQ-Net, DoReFa-Net, and
others. In this case, the computation of an inner product of two $\{n,m\}$-bit
quantized inputs can be computed by a sequence of $n \times m$ binary inner
products. On many common CPUs (including many popular mobile ARMv7 CPUs), this
binary inner product can be efficiently computed by the combination of bitwise
exclusive or (\texttt{xor}) and population count (\texttt{popcnt}) instructions.

In this work, we seek to:

\begin{itemize}
\item Explore the use of these low precision models in the context of mobile
  computer vision applications.
\item Provide the first open-source ARMv7 (and AVX2) runtime for low precision
  neural networks that achieves >80\% of peak theoretical efficiency, using a
  variety of novel tricks (to our knowledge) in the implementation of binary
  convolution kernels.
\item Characterize the performance of this runtimes on a range of mobile CPU
  architectures against state of the art \texttt{float32} and \texttt{int8}
  baselines.
\end{itemize}

We note that there is no single answer to the question ``are ultra-low-precision
models beneficial?''. To decompose the question, we consider the following three
major factors:

\begin{itemize}
\item \textbf{Baseline}: we found that with different baseline choices the
  conclusion may differ much, and as a result a highly optimized baseline
  provides the most fair conclusion on the potential of low precision math.
\item \textbf{Model}: different model parameters (e.g. convolution kernel sizes,
  input sizes) call for different underlying implementations. Sweeping over
  common parameter settings helps in co-designing models and implementations.
  Similar observations have been found in \citet{Vasilache:2014aa}.
\item \textbf{Hardware}: the level of advantage of ultra-low-precision
  implementations differs across CPU microarchitectures. We mainly compare the
  Cortex-A53 and Cortex-A7 chips to exhibit this --- other architectures can be
  evaluated in a similar way.
\end{itemize}

We note that despite a lot of interest being devoted to low-precision machine
learning in recent years, the computational and implementation aspects of the
problem have been relatively under-explored. This is a major gap we seek to
address in our work --- to provide a comprehensive analysis and practical baselines
for future low precision ML applications.

\subsection{Low bitwidth arithmetic}

Consider a standard inner product of two N-bit vectors $x, y$ --- interpreted as
elements of $\{-1, 1\}^N$. Then we have $x \cdot y = N - 2 \cdot
\mathtt{popcnt}(\mathtt{xnor}(x, y))$. This can be generalized to the case where
each element represents an $m$-bit number $x = \sum_{k=0}^m 2^{k} x_{k}$, by
accumulating over each pairwise bit-depth, i.e.

\begin{align*}
  x \cdot y = \sum_{k=0}^m \sum_{j=0}^p 2^{k+j} x_{k} \cdot y_{j} \
\end{align*}

This inner product primitive can be easily generalized to matrix
multiplications, convolutions (via direct methods or Topelitz matrix lowering),
which allows us to build the full set of primitives used for compute-heavy
kernels in DNN workloads. This additionally suggests an BLIS-style
\citep{Van-Zee:2015aa} implementation strategy for these techniques: for each
architecture we target, we can just implement an efficient packed layout (where
we store each bit-depth separately as a binary vector) and a subsequent binary
inner product microkernel, and use generic routines to accumulate partial inner
products, handle tiling, parallelism, and so on.

\subsection{Training low bitwidth convolutional models}

While not a focus of our work, we note that progress in training methods for low
bitwidth models have substantially progressed over the past few months. In
particular, we wish to highlight two techniques which we successfully used for a
number of our models --- the HWGQ quantizer \citep{Cai:2017aa}, and the
bit-decay method for retraining \citep{Wen:2016aa}.

The HWGQ quantizer provides a theoretically optimal quantizer (via an EM
algorithm) for N-bit quantization of a half-Gaussian distribution --- which is
the approximate distribution of $Y \sim \mathtt{ReLU}(\mathtt{BatchNorm}(X))$.
The HWGQ quantizer achieved state of the art results for low-bitwidth training
on a number of modern CNN architectures, including ResNet-50. We find the HWGQ
quantization approach for \textnormal{activations (which is simply a uniform
  quantizer on $[\mathtt{offset}, \mathtt{offset} + \left(2^{bit} - 1 \right) *
  \mathtt{scale}]$) to perform well in our experiments.}

The bit-decay method \citep{Wen:2016aa} is another simple, intuitive trick,
which we found to work well across a range of models. The bit-decay method
trains a series of models at steadily decreasing bit-depths (e.g. 32 bit
$\rightarrow$ 4 bit $\rightarrow$ 2 bit), attempting to ameliorate both the accuracy
loss induced by either aggressive quantization and the optimization challenges
from training these models from scratch. We empirically noticed on small-scale
CIFAR10 experiments that training some models (e.g. ResNet-32 in 2b/2b, as in
HWGQ), we found if difficult to recover the expected \texttt{float32} accuracy.
Using a simple modification to the bit-decay approach --- preferentially
decaying activations --- we were able to recover the accuracy loss in a range of
challenging scenarios such as parameter-poor architectures like MobileNet.

\section{Our binary inner-product microkernel}

A lot of interest in low bitwidth models naturally come from low power and
embedded inference scenarios, which is our area of focus here. We'll first
analyze the theoretical throughput, and describe some tricks we used to achieve
close to peak throughput in our implementation.

A Cortex-A7 is a typical low-end mobile phone CPU, used in hundreds of millions
of mobile devices today (over 25\% of devices using our application). The
relevant inner loops are \texttt{VMLA\.F32} (\texttt{float32}), \texttt{VMULL.U8
  + VPADAL.U16}, (\texttt{int8} --- this can be improved by accumulating intermediates in 8/16 bit precision), and \texttt{VEOR + VCNT.8} (binary inner
products). These have a theoretical throughput of 2 FLOPs/cycle
(\texttt{float32}), 2.5 8-bit OPs/cycle (\texttt{int8}), and 42 binary
OPs/cycle (binary inner products). Similarly, we obtain 8 FLOPs/cycle, 5.3 8-bit
OPs/cycle, and 85 binary OPs/cycle, respectively, on a Cortex-A53. This
indicates the theoretical potential improvements (10--16x) for compute-bound
workloads via low-bitwidth kernels. We achieve these improvements in several
real-world use-cases, as detailed below.

We note a few useful tricks we used in the implementation of our microkernel,
which is a GESS-style microkernel. The challenge is to efficiently implement a
function, taking $A \in \{0, 1\}^{M_T \times K}$ and $B \in \{0, 1\}^{N_T \times
  K}$ (possibly packed) matrices, and computing the matrix product $A \cdot B^T
\in \mathbb{R}^{M_T \times N_T}$, where $M_T, N_T$ are our tile sizes. These
included:

\begin{itemize}
\item The inputs are packed to SIMD width (i.e. 128 bits on ARM NEON), so all
  loads in the microkernel can be computed via the post-increment addressing
  mode on ARMv7, and we avoid any shuffles or masks in our inner loop.
\item We leverage the fact that our accumulation size $K$ is effectively always
  less than $2^{16}$, and commonly less than $2^{10}$. This allows us (via
  modifying our packing and accumulation logic) to maintain multiple parallel
  accumulators while still keeping a relatively large reduction dimension.
\item Fusing our quantization and packing steps (for our kernel variants that
  accept a floating point).
\item Using standard blocking techniques (specifically using a large register
  block, and secondarily L1 cache blocking over LHS and RHS). We note that our
  packed tensors can fit entirely in the L2 cache on many of our
  models and architectures.
\end{itemize}

In microbenchmarks, our microkernel hits a peak of approximately 75\% of peak
performance on these devices. Our code is open-sourced at
\url{http://anonymous.url}.

\section{Performance Experiments}

\subsection{Baseline implementations}

We benchmark on both a 1.2GHz Cortex-A7 (Qualcomm Snapdragon 200) and a 1.4GHz
Cortex-A53 (Qualcomm Snapdragon 410) CPU, two common and relatively low-end
mobile chipsets, with GCC 4.9 targeting ARMv7.

We rely on two libraries for current state-of-the-art convolution performance
baselines --- GEMMLOWP \citep{Google:aa} (for
\texttt{int8} convolutions), and
NNPACK \citep{Dukhan:aa} (for
\texttt{float32} convolutions). For a given benchmark configuration of GEMMLOWP
and NNPACK, we report the highest performing variant. Our $3 \times 3$
convolution implementation is a variant of the $F(6 \times 6, 3 \times 3)$
Winograd transform approach \citep{Lavin:2016aa}, with a few modifications ---
we cache the transformed filter weights, (which avoids repeated computation at
inference time, in exchange for increased memory footprint), and we optionally
store intermediate transformed data in \texttt{float16} (which trades off a
reduction precision by reducing the memory bandwidth). These changes improve
performance by on the order of 20\% on common convolution layers compared to the
standard NNPACK baseline, and improve performance by up to 2.5x compared to the
GEMMLOWP Topelitz lowering approach.

Depthwise convolutions have become increasingly popular for low-flop CNN
architectures \citep{Howard:2017aa,Zhang:2017aa,Zoph:2017aa}. While we don't
focus on these here, we note that their low arithmetic intensity makes these
convolutions quite amenable to ultra-low-precision implementations.

In parallel with our work, we note that \citep{Umuroglu:2017aa} pursued a
similar line of work. We incorporated their code, but found it uniformly
substantially slower that our implementation (the geometric mean speedup over
parameters of interest was 3.7x). We suspect this is due to the authors not
focusing on ARMv7. We note that our baseline is significantly improved in many
cases.

\subsection{Results}

We examine the peak performance of our micro-kernels, and overall performance on
convolution and matrix multiplication sizes used in modern deep learning
applications. That is, we focus on batch-size 1 (as is common in inference
workloads) $1 \times 1$ and $3 \times 3$ convolutions, with varying spatial
dimensions $S \in \{14, 28, 56, 104\}$ and input/output channels $C \in \{64,
128, 256, 384, 512, 768, 1024\}$. We report single-threaded runtimes (although
multi-threading leads to very similar conclusions).The ratio of binary GOP/s to
\texttt{max}(GEMMLOWP GOP/s, NNPACK GFLOP/s) is reported at each point we sample
in ($C$, $S$) space.

\begin{figure}[h]
  \centering
  \scalebox{0.8}{\input{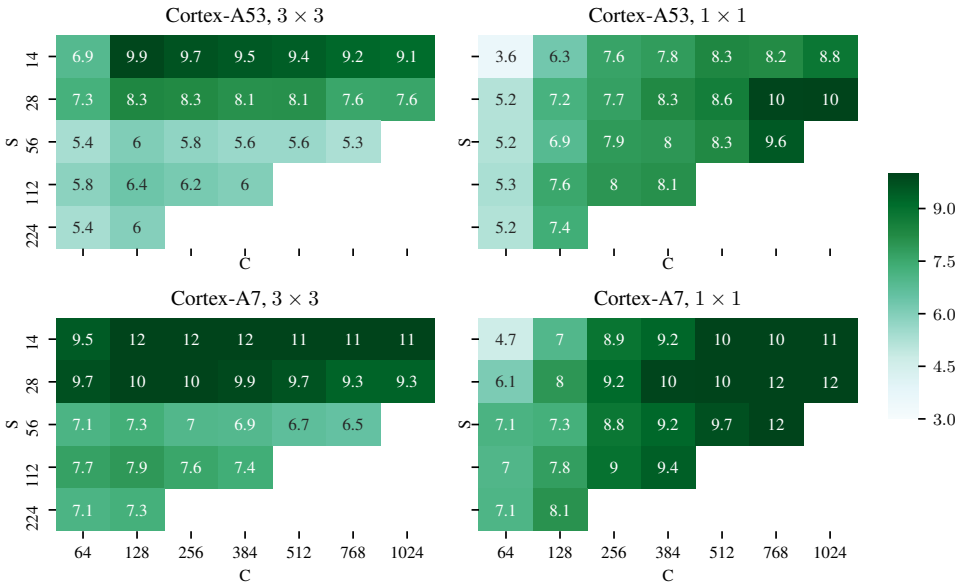}}
  \caption{Performance improvement (the ratio of binary GOP/s over the best
    baseline GOP/s) for $3 \times 3$ and $1 \times 1$ convolutions on
    Cortex-A7 and Cortex-A53, for a given spatial input $S \times S$ with input
    and output channels $C$.}
\end{figure}

We see that for both microarchitectures, we can achieve substantial speedups
with our ultra-low-precision convolution implementations for a range of
achievable bitwidths. These results are, to our knowledge, the state of art for
ARMv7, substantially outperforming existing open-source implementations
\citep{Umuroglu:2017aa,Yang:2017aa}, by more than an order of magnitude in some
cases.

Our results show that we can compute approximately 9 binary convolutions (e.g.
using 3 bit activations and weights) and still achieve performance improvements
compared to existing well-tuned baselines. With recently developed techniques
such as HWGQ-Net (2 bit activations, 1 bit weights) and retraining with
bit-decay (2 bit activations, 2 bit weights), these aggressively quantized
models are becoming more and more achievable. Our work described here makes
these models practically applicable on a huge number of ARMv7 devices used
today.

\small

\bibliography{main}

\end{document}